\documentclass[10pt,twocolumn,letterpaper]{article}
\usepackage{cvpr}

\usepackage{graphicx}
\usepackage{amsmath}
\usepackage{amssymb}
\usepackage{booktabs}
\usepackage{times}
\usepackage{microtype}
\usepackage{epsfig}
\usepackage[table,xcdraw]{xcolor}
\usepackage{caption}
\usepackage{float}
\usepackage{placeins}
\usepackage{color, colortbl}
\usepackage{stfloats}
\usepackage{tabularx}
\usepackage{xstring}
\usepackage{multirow}
\usepackage{xspace}
\usepackage{url}
\usepackage{subcaption}
\usepackage{xcolor}
\usepackage[hang,flushmargin]{footmisc}
\usepackage{gensymb}
\usepackage{xr-hyper}

\renewcommand{\paragraph}[1]{
    \vspace{0.5mm}
     \noindent\textbf{#1.} 
}

\usepackage[pagebackref,breaklinks,colorlinks]{hyperref}
\usepackage[capitalize]{cleveref}
\crefname{section}{Sec.}{Secs.}
\crefname{table}{Table}{Tables}
\crefname{figure}{Fig.}{Figs.}

\graphicspath{ {images/} }

\title{Street-View Image Generation from a Bird's-Eye View Layout}

\author{Alexander Swerdlow\qquad Runsheng Xu\qquad Bolei Zhou \\University of California, Los Angeles\\{\tt\small \{aswerdlow, rxx3386\}@ucla.edu, bolei@cs.ucla.edu}}

\begin{document}

\maketitle
\thispagestyle{empty}
\pagestyle{empty}

\begin{abstract}
  Bird's-Eye View (BEV) Perception has received increasing attention in recent years as it provides a concise and unified spatial representation across views and benefits a diverse set of downstream driving applications. At the same time, data-driven simulation for autonomous driving has been a focal point of recent research but with few approaches that are both fully data-driven and controllable. Instead of using perception data from real-life scenarios, an ideal model for simulation would generate realistic street-view images that align with a given HD map and traffic layout, a task that is critical for visualizing complex traffic scenarios and developing robust perception models for autonomous driving. In this paper, we propose BEVGen, a conditional generative model that synthesizes a set of realistic and spatially consistent surrounding images that match the BEV layout of a traffic scenario. BEVGen incorporates a novel cross-view transformation with spatial attention design which learns the relationship between cameras and map views to ensure their consistency. We evaluate the proposed model on the challenging NuScenes and Argoverse 2 datasets. After training, BEVGen can accurately render road and lane lines, as well as generate traffic scenes with diverse different weather conditions and times of day. The code is open-source and available at \href{https://metadriverse.github.io/bevgen/}{metadriverse.github.io/bevgen}.
\end{abstract}

\section{Introduction}
\label{sec:intro}

BEV perception for autonomous driving is a fast-growing research area, with the goal of learning a cross-view representation that transforms information between a perspective and a bird's-eye view. Such a representation can be used in downstream tasks such as path planning and trajectory forecasting~\cite{zhangBEVerseUnifiedPerception2022}. The recent successes in BEV perception, whether for monocular~\cite{gosalaBirdSEyeViewPanoptic2022} or multi-view images~\cite{zhouCrossviewTransformersRealtime2022, xu2022cobevt}, focus on the predictive aspect of BEV perception with street-view images as input and a semantic BEV layout as output. However, the generative side of BEV perception—which aims at synthesizing realistic street-view images from a given BEV semantic layout—is far less explored. A BEV layout concisely describes a traffic scenario at the semantic level and thus it is a natural representation to use to generate corresponding street-view images. This is the first work we are aware of to introduce and tackle such a task. There are many applications for this new BEV-conditioned generative model. For example, we can create synthetic training data for perception models or visualize safety-critical situations. 

\begin{figure}[t]
  \parindent=8.5pt{\small BEV Layout}\hspace{1.4cm} \small{Generated Street-View Images} \par\medskip
  \centering
   \includegraphics[width=1.004\columnwidth]{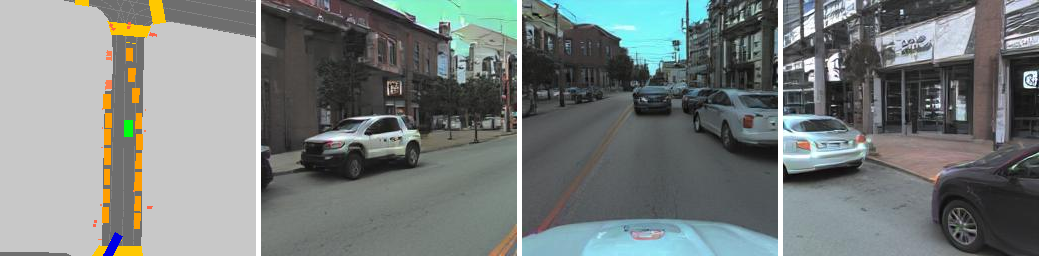}
  \vfil
  \vspace{0.05cm}
   \includegraphics[width=1.004\columnwidth]{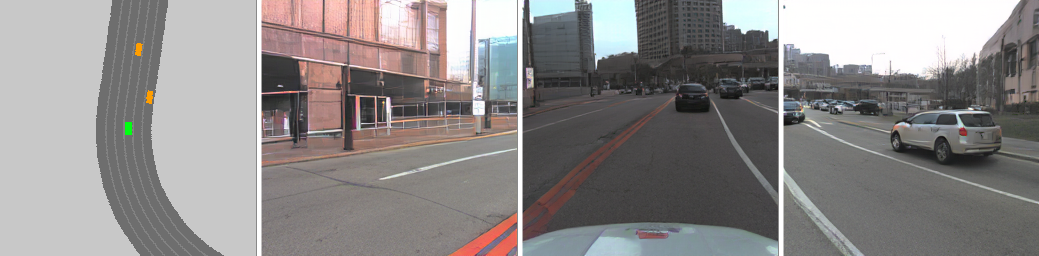}
   \caption{The proposed BEVGen model generates realistic and spatially consistent street-view images from BEV layout. There are 3 camera views surrounding the ego vehicle as indicated by the green rectangle in the BEV layout.}
   \label{fig:main_images}
\end{figure}

Whereas most current approaches to synthetic training data involve a complex simulator or 3D reconstructed meshes, a controllable generative model is not only simpler but naturally provides diverse image generation. Another benefit provided by BEV generation is the ease of visualizing and editing traffic scenes. In the case of self-driving vehicles, we often care about a small set of rare scenarios where an accident is likely to happen. These corner-cases represent the long-tail distribution that is a key obstacle to robust autonomous driving~\cite{makansiExposingChallengingLong2021,breitensteinCornerCasesVisual2021}. Human users can intuitively edit a BEV layout and then use a generative model to output a diverse set of corresponding street-view images for the stress test of the driving system. As shown in Fig.\ref{fig:main_images}, our proposed \textbf{BEVGen} model can synthesize realistic street-view images from the BEV layout either collected from real world or provided by a driving simulator.

The fundamental question for BEV generation is: what is a plausible set of street-view images that correspond to a BEV layout? One could think of numerous images with varying vehicle types, backgrounds, and more. For a set of views to be realistic, we must consider several properties of the images. Similar to the problem of novel view synthesis, images must appear consistent—as if they were taken in the same physical location. For instance, cameras with an overlapping field-of-view (FoV) must ensure overlapping content is correctly shown, accounting for the shifted perspective. The visual styling of the scene also needs to be consistent such that all virtual views appear to be created in the same geographical area (e.g., urban vs. rural), at the same time of day, with the same weather conditions, and so on. In addition to this consistency, the images must correspond to the HD map, faithfully reproducing the specified road layout, lane lines, and vehicle locations. Unlike image-to-image translation with a semantic mask—which has significant amount of prior work—a BEV generation model must infer the image layout to account for occlusions between objects and the relative heights of objects in a scene.

In this work, we tackle this new task of generating street-view images from a BEV layout and propose a generative model to address the underlying challenges. We develop \textbf{BEVGen}, an autoregressive neural model that generates a set of $n$ realistic and spatially consistent images as seen in \cref{fig:main_images}. \textbf{BEVGen} has two technical novelties: (i) it incorporates spatial embeddings using camera intrinsics and extrinsics to allow the model to attend to relevant portions of the images and HD map, and (ii) it contains a novel attention bias and decoding scheme that maintains both image consistency and correspondence. Thus the model can generate high-quality scenes with spatial awareness and scene consistency across camera views. Compared to baselines, the proposed model obtains substantial improvement in terms of image synthesis quality and semantic consistency. The model can also render realistic scene images from out-of-domain BEV maps, such as those provided by a driving simulator or edited by a user. We summarize our contributions as follows:

\begin{itemize}
\vspace{0pt}
  \setlength{\itemsep}{0pt}
  \setlength{\parsep}{0pt}
  \setlength{\parskip}{0pt}
  \item We tackle the new task of multi-view image generation from BEV layout. It is the first attempt to explore the generative side of BEV perception for driving scenes.
  \item We develop a novel generative model, \textbf{BEVGen}, that can synthesize spatially consistent street-view images by incorporating spatial embeddings and a pairwise camera bias.
  \item The model achieves high-quality synthesis results and shows promise for applications such as data augmentation and simulated driving scene rendering.
\end{itemize}

\section{Related Work}
\label{sec:related}

Synthetic data-generation for autonomous driving has been widely studied, with early approaches relying on handcrafted assets in a game engine, and more recent data-driven approaches that augment or entirely replace specified handcrafted assets with learned generative models~\cite{mutschModelBasedDataDrivenSimulation2023}. Prior works have demonstrated purely data-driven simulation~\cite{kimDriveGANControllableHighQuality2021,aminiLearningRobustControl2020} but have not tackled multi-view generation and lack the ability to arbitrarily control the position of actors and objects.

Outside of autonomous driving applications, there has also been significant work on image generation models. One large area of research has been on image generation conditioned on modalities such as text and audio~\cite{rameshZeroShotTexttoImageGeneration2021, liDirectSpeechtoimageTranslation2020} or more directly through semantic masks~\cite{isolaImageToImageTranslationConditional2017} or bounding boxes~\cite{liImageSynthesisLayout2021}. Our task has spatial constraints as in the latter tasks, but is distinct in that the constraints must be inferred from an alternate perspective which requires a 3D understanding of the scene.

Next, our task is closely related to—but distinct from, novel view synthesis (NVS), where the goal is to generate an image of a scene from a virtual perspective given a true image from a different perspective. This has been studied in the context of transforming satellite images into a corresponding street-view image, which is commonly referred to as cross-view synthesis~\cite{tokerComingEarthSatellitetoStreet2021}. Other approaches to NVS focus on smaller viewpoint transformations and use warping~\cite{wilesSynSinEndtoEndView2020} or simply implicitly learn the transformation ~\cite{rombachGeometryFreeViewSynthesis2021,renLookOutsideRoom2022}. BEV Generation requires a similar 3D understanding between different viewpoints as in NVS, but lacks the conditioning information provided by a source view(s) and requires consistency not only between frames but also with an HD map.
\section{Method}
\label{sec:method}

In this section, we introduce the framework of the proposed BEVGen. The problem definition of BEV generation is that given a semantic layout in Birds-Eye View (BEV), $B\in\mathbb{R}^{H_b \times W_b \times c_{b}}$ with the ego at the center and $c_b$ channels describing the locations of vehicles, roads, lane lines, and more (see \cref{ssec:dataset}), we would like to generate $n$ images $I_k\in\mathbb{R}^{H_c \times W_c \times 3}$ under a set of $n$ virtual camera views $(K_k, R_k, t_k)_{k=1}^{n}$, where $K_k, R_k, t_k$ are the intrinsics, extrinsic rotation, and translation of the $k$th camera.

\begin{figure*}[t]
 \vspace{-3mm}
  \centering
   \includegraphics[width=1.0\linewidth]{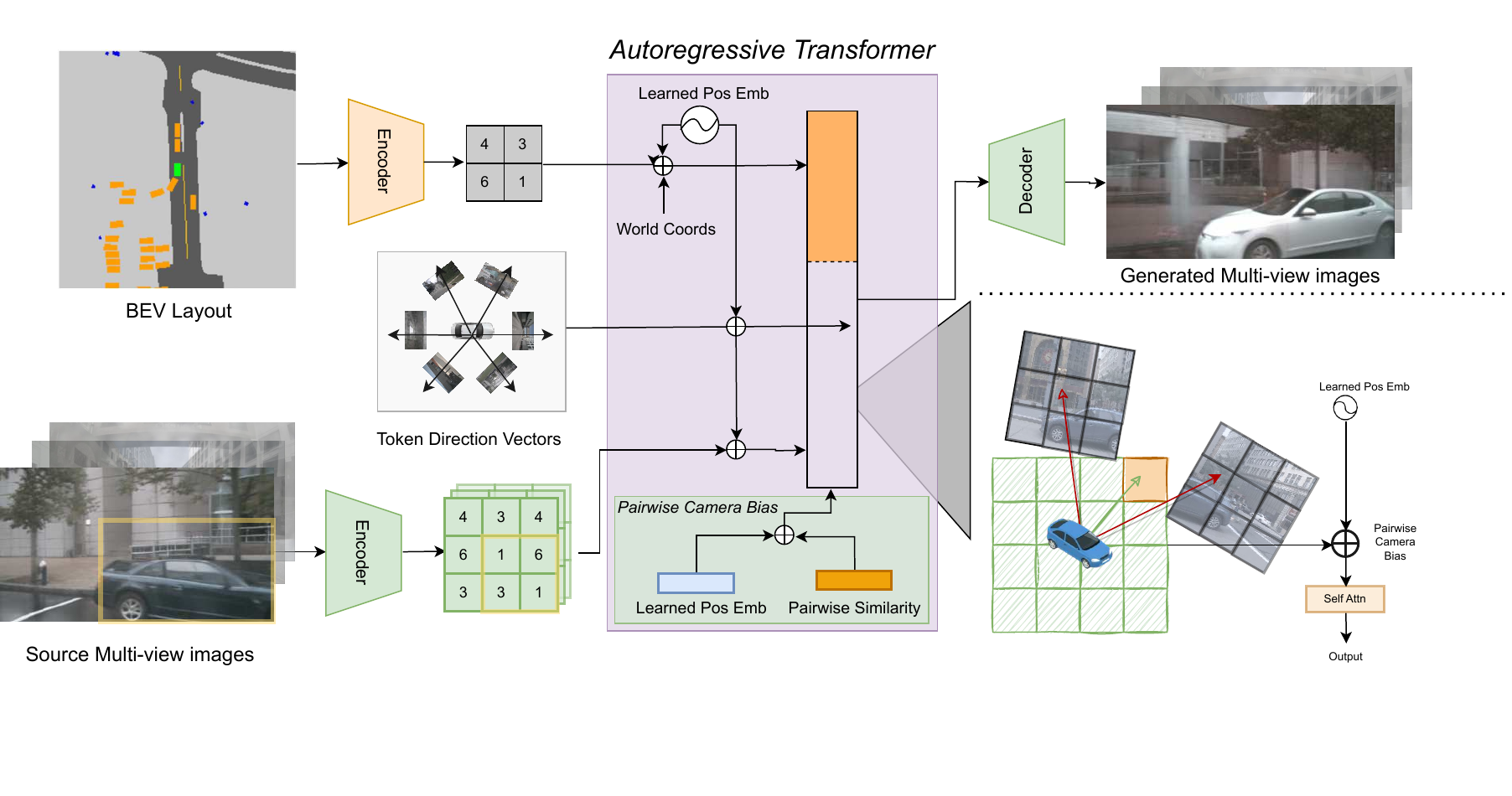}
   \vspace{-14mm}
   \caption{BEVGen framework. A BEV layout and source multi-view images are encoded to a discrete representation and are flattened before passed to the autoregressive transformer. Spatial embeddings are added to both camera and BEV tokens inside each transformed bloc. We construct a pairwise matrix that encodes relationship between a given image token and another BEV/image token (bottom right). This learned pairwise camera bias is added to the attention weights. We pass the generated tokens to the decoder to obtain generated images during inference.}
   \label{fig:model_architecture}
\end{figure*}

\cref{fig:model_architecture} illustrates the framework of the proposed BEVGen. BEVGen consists of two autoencoders modeled by VQ-VAE, one for images and one for the BEV representation, that allow the causal transformer to model scenes at a high level. The key novelty lies in how the transformer can relate information between modalities and across different views. The cross-view transformation encodes a cross-modal inductive 3D bias, allowing the model to attend to relevant portions of the HD map and nearby image tokens. We explain each part in more detail below.

\subsection{Model Structure}

\paragraph{Image Encoder} To generate a globally coherent image, we model our distribution in a discrete latent space instead of pixel space. We use the VQ-VAE model introduced by Oord \textit{et~al.}~\cite{vandenoordNeuralDiscreteRepresentation2017} as an alternative generative architecture to GANs\footnote{Note that switching to the recently developed class of diffusion models can potentially improve the image synthesis quality, but such models require an order of magnitude of additional data and computational resources for training and thus we leave it for future works.}. We replace the $L_2$ with a perceptual loss and incorporate a patch-wise adversarial loss as in~\cite{esserTamingTransformersHighResolution2021}. The VQ-VAE architecture consists of an encoder $E^{c}$, a decoder $D^{c}$, and a codebook $\mathcal{Z}^c=\left\{z_{f}\right\}_{f=1}^{F_c} \subset \mathbb{R}^{n_c}$ where $F_c$ is the number of code vectors and $n_c$ is the embedding dimension of each code. Given a source image, $x_k\in\mathbb{R}^{H_c \times W_c \times 3}$, we encode $\hat{z}^{c}_{k} = E^c(x_k) \in \mathbb{R}^{h_c \times w_c\times n_c}$. To obtain a discrete, tokenized representation, we find the nearest codebook vector for each feature vector $\hat{z}^{c}_{k,ij} \in \mathbb{R}^{n_{c}}$ where $i,j$ are the row, column indices in the discrete latent representation with size $h_c \times w_c$:
\begin{equation}
 z^{c}_{k,ij}=\underset{f}{\arg \min }\left\|\hat{z}^{c}_{k,ij}-z_{f}\right\| \in \mathbb{N}.
  \label{eq:codebook}
\end{equation}
This creates a set of tokens $z^{c}_k \in \mathbb{N}^{h_c \times w_c}$ that we refer to as our image tokens. To generate an image from a set of tokens, we first obtain $\tilde{z}^{c}_k\in\mathbb{R}^{h_c \times w_c \times n_c}$, by using $z_k^c$ as an index to our codebook, $\mathcal{Z}_c$. We can later decode with a convolutional decoder, $D^{c}(\tilde{z}^{c}_k)\in\mathbb{R}^{H_c \times W_c \times 3}$, using the same architecture as~\cite{esserTamingTransformersHighResolution2021}.

\paragraph{BEV Encoder} To condition our model on a BEV layout, we use a similar discrete representation as for camera images, except we replace the perceptual and adversarial losses with a binary cross entropy loss for binary channels and an $L_2$ loss for continuous channels. We encode our BEV map $h$ as before with $E^{b}(h) \in \mathbb{R}^{h_b \times w_b\times n_b}$ and $\mathcal{Z}^b=\left\{z_{f}\right\}_{f=1}^{F_b} \subset \mathbb{R}^{n_b}$ to obtain a set of tokens, $z^{b} \in \mathbb{N}^{h_b \times w_b}$. We discard the decoder stage, $D^b$, after training the 1st stage as it is not needed for our second stage or for inference.

\paragraph{Autoregressive Modeling}
\label{ssec:autoregressive}
Given a BEV layout and a set of $n$ camera parameters, we seek to generate $n$ images by learning the prior distribution of a set of discrete tokens, $z^c$ conditioned on $z^{b}, (K_k, R_k, t_k)_{k=1}^{n}$.
\begin{equation}
 p(z^c|z^{b}, K, R, t) = \prod_{i = 0}^{h_c \times w_c \times n} p(z^c_i | z^{c}_{<i}, z^{b}, K, R, t).
  \label{eq:autoregressive}
\end{equation}
We model $p(.)$ by training a transformer $\tau$ that iteratively predicts a probability distribution over all possible tokens based on prior image tokens, discretized BEV features, and their respective camera parameters. We choose a transformer architecture as it provides global attention which aids in cross-view consistency. To implement this, we perform causal self-attention between image tokens such that image tokens cannot look at future tokens in the attention process. We allow global attention between image tokens and all BEV tokens. This serves as an extension to the prior modeling proposed in~\cite{vandenoordNeuralDiscreteRepresentation2017,esserTamingTransformersHighResolution2021} which we refer to for further details.

\subsection{Spatial Embeddings}
\label{ssec:spatial_embedding}
To help the model attend to relevant tokens both in the camera and BEV feature space, we introduce positional embeddings. We take inspiration from work on BEV segmentation~\cite{zhouCrossviewTransformersRealtime2022} on alignment between the BEV and first-person view (FPV) perspectives.

\paragraph{Camera Embedding} In order to align image tokens with BEV tokens, we use the known intrinsics and extrinsics to reproject from image coordinates to world coordinates. Given a token in image space, $z^{c}_{k,ij}\in\mathbb{R}^2$, we convert to homogeneous coordinates, $\bar{z}^{c}_{k,ij}$ and obtain a direction vector in the ego frame as follows:

\begin{align}
d_{k,ij} = R^{-1}_k K^{-1}_k \bar{z}^{c}_{i,jk} - t_k
\label{eq:directionvector}
\end{align}

We use a 1D convolution, $\theta_c(d)\in \mathbb{R}^{n \times h_c \times w_c \times n_{emb}}$, to encode our direction vector in the latent space of the transformer, $n_{emb}$. We encode our image tokens using a shared learnable embedding $\lambda_c(z^{c}_{k, ij}) \in \mathbb{R}^{n_{emb}}$, and add a per-token learnable parameter, $\Lambda^{c}_{k, ij}\in \mathbb{R}^{n_{emb}}$, across image tokens:
\begin{equation}
 l_{k,ij} = \lambda(z^{c}_{k, ij}) + \theta(d_{k,ij}) + \Lambda_{k, ij}.
  \label{eq:image_embedding}
\end{equation}
\paragraph{BEV Embedding} To align our BEV tokens with our image tokens, we perform a similar operation as in \cref{eq:image_embedding} and use the known BEV layout dimensions to obtain coordinates in the ego frame. We obtain $t_{yx}\in \mathbb{R}^2$, where $y,x$ correspond to the row and column indices of discrete BEV representation, for each token and encode these into our transformer latent space, with $\theta_b(t)\in \mathbb{R}^{h_b \times w_b \times n_{emb}}$. We similarly use a shared learnable embedding for our discrete tokens, $\lambda(z^{b}_{yx}) \in \mathbb{R}^{n_{emb}}$, and a per-token learnable parameter, $\Lambda_{yx}$:

\begin{equation}
 l_{yx} = \lambda(z^{b}_{yx}) + \theta_b(t_{yx}) + \Lambda_{yx}.
  \label{eq:bev_embedding}
\end{equation}

where $l$ represents the final input embeddings for the transformer
decoder block.

\subsection{Camera Bias}
\label{ssec:camera_bias}
In addition to providing the model with aligned embeddings, we add a bias to our self-attention layers that provides both an intramodal (image to image) and intermodal (image to BEV) similarity constraint. This draws inspiration from~\cite{renLookOutsideRoom2022}, but instead of providing a blockwise similarity matrix that is composed of encoded poses between frames, we provide a per-token similarity based on their relative direction vectors. Our approach also encodes the relationship between image and BEV tokens. For self-attention in any layer between some query $q_r \in \mathbb{R}^{n_{emd}}$ and key/value $k_c$/$v_c \in \mathbb{R}^{n_{emd}}$, where $r, c$ are the row, column of the pairwise attention matrix, we have:
\begin{equation}
 \text{Attention}(q_r, k_c, v_c)=v_c\operatorname{softmax}\left(\frac{a_{rc}}{\sqrt{d}}\right),
 \label{eq:attention}
\end{equation}
\begin{equation}
 a_{rc} = q_r k_c + \beta_{rc}.
 \label{eq:weight}
\end{equation}

 The transformer sequence is composed of $h_b w_b$ conditional BEV tokens followed by camera tokens. If $r,c > h^{b}w^{b}$, both positions correspond to image tokens and thus we have two direction vectors, $d_r, d_c$, computed as in \cref{eq:directionvector}. As discussed in \cref{ssec:autoregressive}, we have a mapping between the sequence index and image token $(i,j)$ in camera $k$. If $r > h^{b}w^{b} > c$, we have a query for some image token and a key/value pair corresponding to BEV token. Thus, we again construct two direction vectors. In this case our BEV direction vector consists of the 2D World coordinates (in the ego-center frame) and our image direction vector is the same as in \cref{eq:directionvector} except with the row value as the center of the image. Given these two direction vectors, $d_r, d_c$, we add the cosine similarity and a learnable parameter, $\theta_{rc}$, as shown in the bottom right of \cref{fig:model_architecture}: 

\begin{equation}
  \beta_{rc} = \frac{d_r \cdot d_c}{\|d_r\| \|d_c\|} + \theta_{rc}.
  \label{eq:attenion_bias}
 \end{equation}
\section{Experiments}
\label{sec:experiments}

\subsection{Datasets}
\label{ssec:dataset}
 We evaluate the proposed method using the nuScenes dataset~\cite{caesarNuScenesMultimodalDataset2020} and Argoverse 2 dataset~\cite{wilsonArgoverseNextGeneration2021}, which are commonly used for BEV segmentation and detection. Each instance in both of our datasets contain ground-truth 3D bounding boxes, mutli-view camera images, calibrated camera intrinsics and extrinsics, and LiDAR. We project these 3D bounding boxes onto a BEV layout following standard practice used in BEV segmentation~\cite{zhouCrossviewTransformersRealtime2022,philionLiftSplatShoot2020}.

\paragraph{nuScenes} The nuScenes dataset consists of 1000, 20-second scenes captured at 12Hz by 6 cameras, which provide full $360\degree$ camera coverage. However, synchronization between cameras occurs only at 2Hz so we increase our dataset size by re-sampling at 20Hz, pruning instances where any camera is more than 100ms outside of the frame. We linearly interpolate from the nearest ground-truth annotations to create paired annotation data. This provides roughly 9x the number of instances as the original training set, for a total of 260k instances. For validation, we use the standard set containing 6k instances. For all visualizations, we flip the back left and right cameras along the vertical axis to highlight the image consistency of our model.

\paragraph{Argoverse 2}
Argoverse 2 comprises 1000, 15-second scenes annotated at 10hz, with a synchronized camera captured at 20hz. Compared to Argoverse 1~\cite{changArgoverse3DTracking2019}, it covers much broader and more diverse image content and is comparable to the size of nuScenes dataset. This provides 210k and 30k instances for train and validation, respectively. As the front camera is rotated $90\degree$, we extract a square crop from the front three cameras for all experiments.

\paragraph{Preprocessing} The BEV layout representation used in training and testing covers $80m \times 80m$ around the ego center. On nuScenes, there are 21 channels, with 14 channels being binary masks representing map information (lane lines, dividers, etc.) and actor annotations (cars, trucks, pedestrians, etc.). The remaining 7 channels provide instance information, including the visibility and size of the annotation. On nuScenes, we resize all images to $H \times W = 224 \times 400$, resulting in a discrete latent representation of $h_c \times w_c = 14 \times 25$. On Argoverse 2, we crop to $H \times W = 256 \times 256$, resulting in $h_c \times w_c = 16 \times 16$. Our BEV layout has a discrete latent representation of $h_b \times w_b = 16 \times 16$. We appropriately modify intrinsics after cropping and resizing. To enable our weighted cross-entropy loss, we project the provided 3D annotations onto the camera frame and weight the corresponding tokens in our discrete camera frame representation, $z_k \in \mathbb{N}^{h_c \times w_c}$.

\subsection{Training Details}

\paragraph{VQ-VAE} We train the 1st stage camera VQ-VAE with aggressive augmentation consisting of flips, rotations, color shifts, and crops. Similarly, we train our 1st stage BEV VQ-VAE with flips and rotations. For the 2nd stage, we add minimal rotations and scaling, as well as cropping. 

\paragraph{Transformer} Our transformer is GPT-like~\cite{radfordLanguageModelsAre} with 16-heads and 24-layers. We use DeepSpeed to facilitate sparse self-attention and 16-bit training. We clip gradients at 50 and use the AdamW optimizer~\cite{loshchilovDecoupledWeightDecay2019} with $\beta_1, \beta_2 = 0.9, 0.95$ and a learning rate of $\lambda=5\text{e-}7$. Both the BEV and image codebooks have $|\mathcal{Z}_c| = |\mathcal{Z}_b| = 1024$ codes with an embedding dimension, $n_c = n_b = 256$.

Additionally, we develop a lighter-weight model that uses sparse attention as in~\cite{zaheerBigBirdTransformers2020}. As opposed to a uniform random attention mask, we unmask regions of the image near the token we attend. Using the same formulation as in~\cref{eq:attenion_bias}, we create a pairwise similarity matrix for image tokens only. As sparse attention groups the input sequence into discrete blocks, we perform average pooling on these blocks and use these values as weights for sampling. Additionally, we have a sliding window in which we always attend to the last $r$ tokens, and we attend to all BEV tokens. For our sparse models, we have an attention mask density of 35\% with a sliding window length of $r = 96$. Except as described in \cref{ssec:ablation}, our sparse model is derived from fine-tuning our full-attention model for 10 epochs.

\subsection{Results} 
\label{ssec:results}
\begin{table}
  \centering
\setlength{\tabcolsep}{4.0pt}
  \begin{tabular}{@{}lcccc@{}}
    \toprule
    Method & FID$\downarrow$ & Road mIoU$\uparrow$ & Vehicle mIoU$\uparrow$\\     
            \midrule
            X-Seq\cite{regmiCrossViewImageSynthesis2018} & 138.30 & - & - \\
            \textbf{BEVGen} & 25.54 & 50.20 & 5.89 \\
            \textbf{BEVGen} Sparse & 28.67 & 50.92 & 6.69\\
    \bottomrule
  \end{tabular}
  \caption{Baseline Comparison over all 6 views on nuScenes Validation.}
  \label{tab:baselines}
\end{table}

\begin{figure*}[t]
   \centering
   \includegraphics[width=0.495\linewidth]{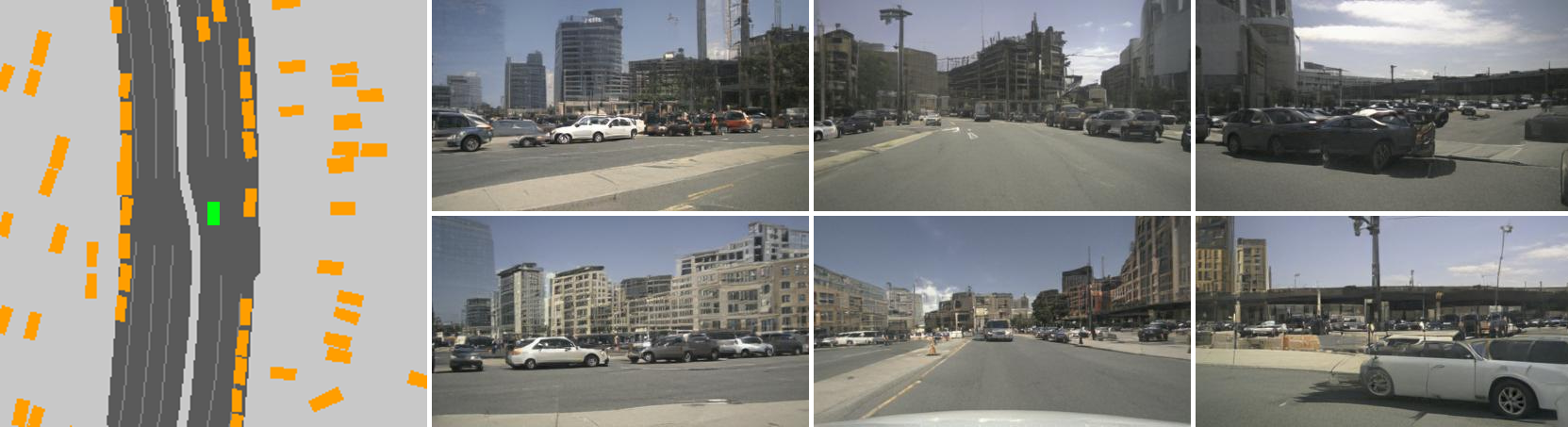}\hfil
   \includegraphics[width=0.495\linewidth]{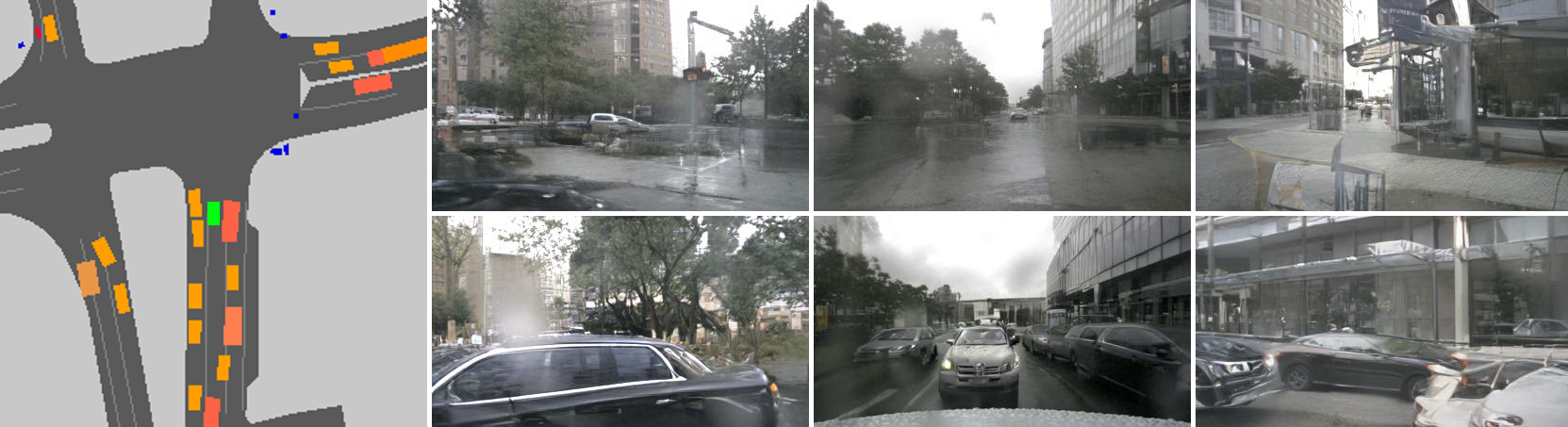}
   \vfil
   
   \centering
   \vspace{0.05cm}
   \includegraphics[width=0.495\linewidth]{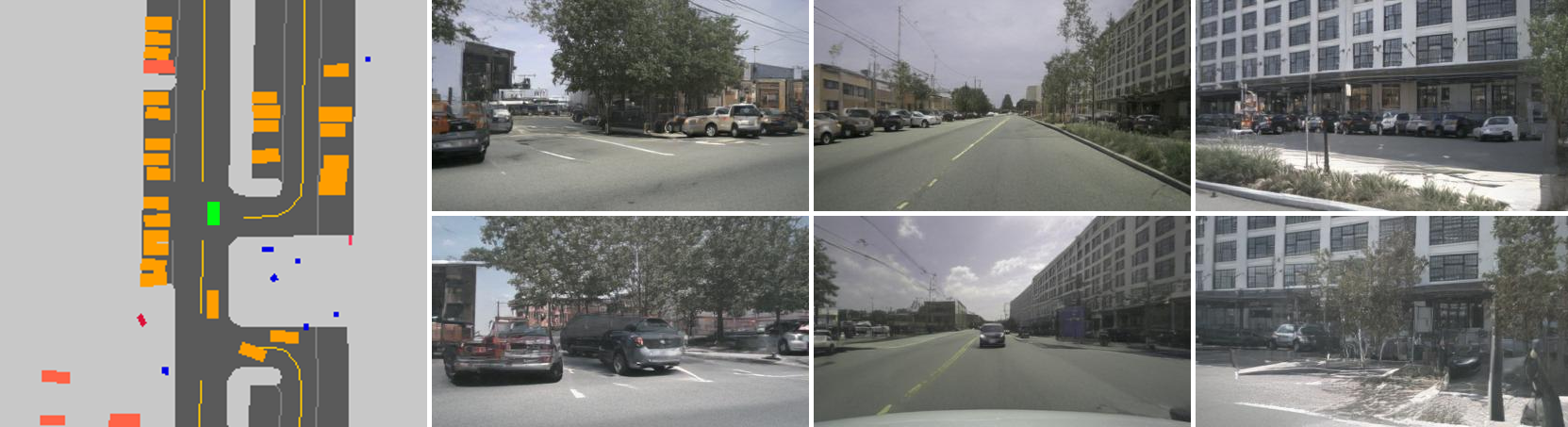}\hfil
   \includegraphics[width=0.495\linewidth]{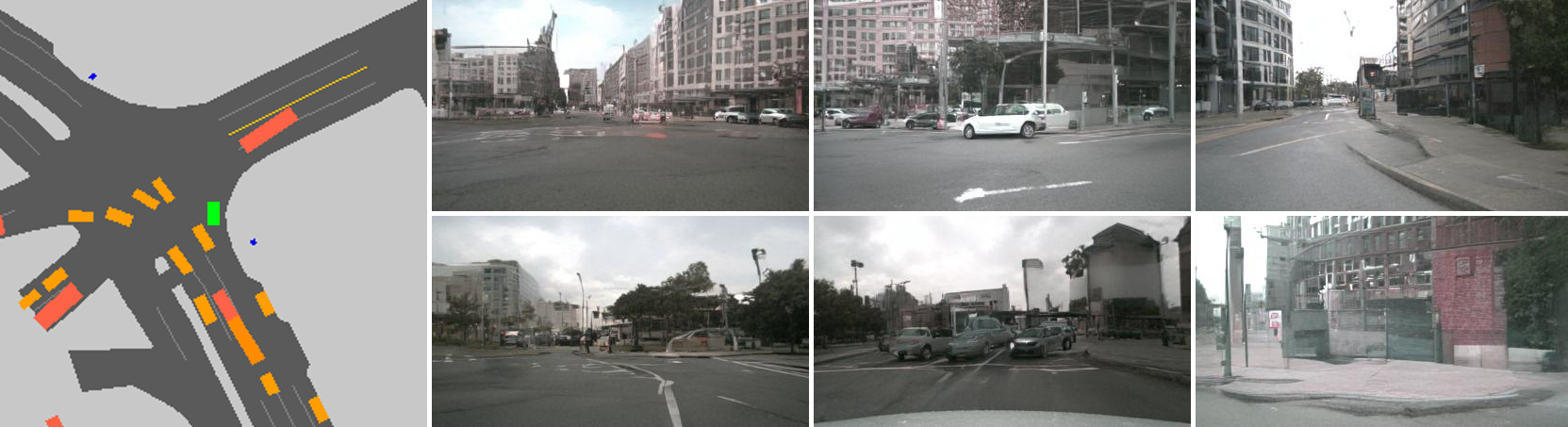}
   \vfil
   
   \centering
   \vspace{0.05cm}
   \includegraphics[width=0.495\linewidth]{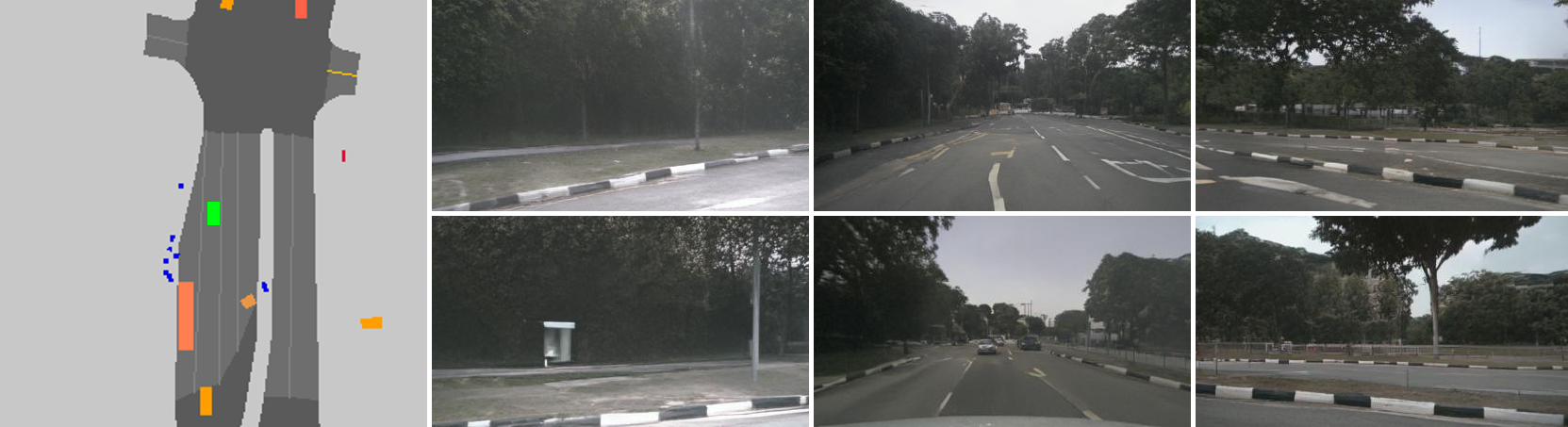}\hfil
   \includegraphics[width=0.495\linewidth]{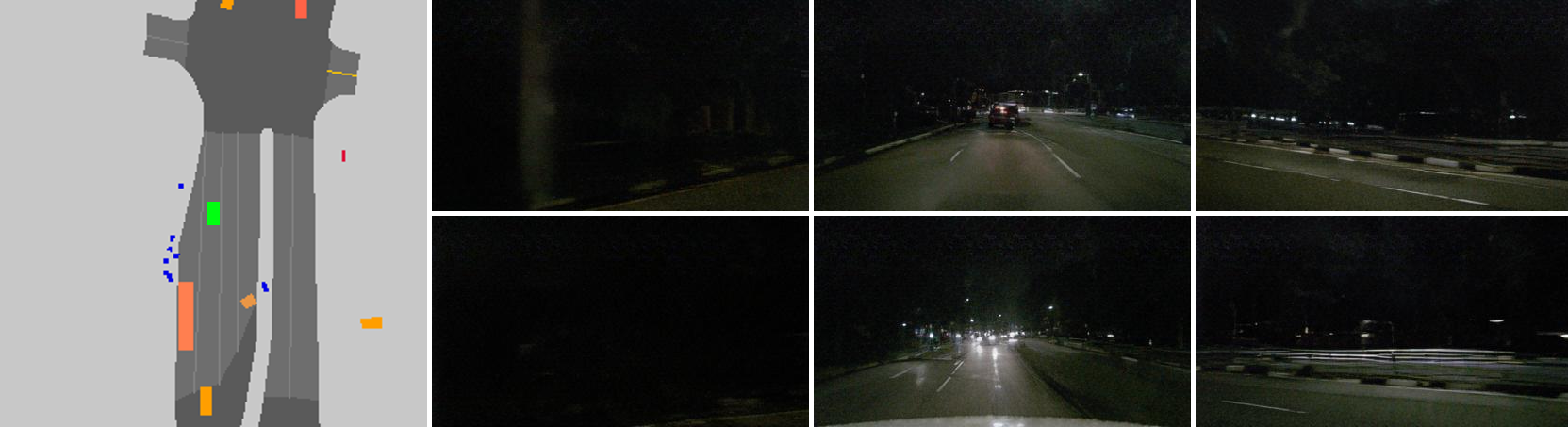}
   
   \caption{Synthesized multi-view images from BEVGen on nuScenes. Image contents are diverse and realistic. The two instances in the bottom row use the same BEV layout for synthesizing the same location in day and night.}
   \label{fig:random_gen}
   
\end{figure*}

\paragraph{Qualitative result}
\cref{fig:random_gen} exhibits the generation examples from \textbf{BEVGen} on nuScenes. Our model is able to generate a diverse set of scenes including intersections, parking lots, and boulevards. We observe that each camera view not only correctly displays the surrounding of the same location, but also preserves the spatial perspective. \textbf{BEVGen} synthesizes images under various weather conditions, with the same weather apparent in all images, including physical artifacts such as rain. We also demonstrate that our model is capable of generating diverse scenes corresponding to the same BEV layout. We see at the bottom of \cref{fig:random_gen} the same location rendered in the day and at night by the model.

\begin{figure*}[t]
  \centering
   \includegraphics[width=1.0\linewidth]{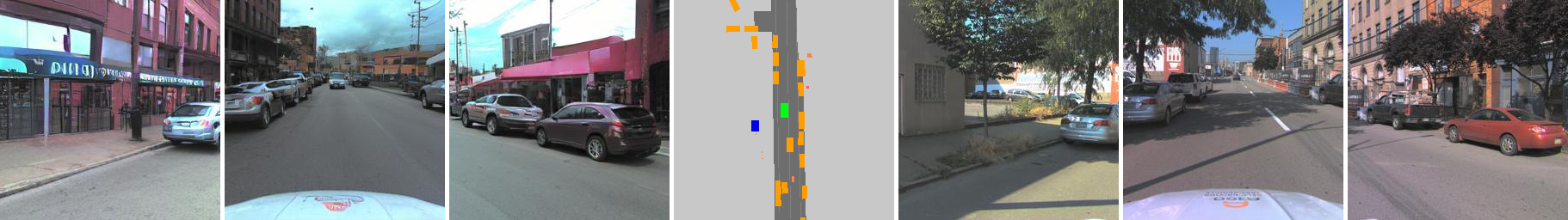}
    \vfil
   \centering
   \vspace{0.05cm}
    \includegraphics[width=1.0\linewidth]{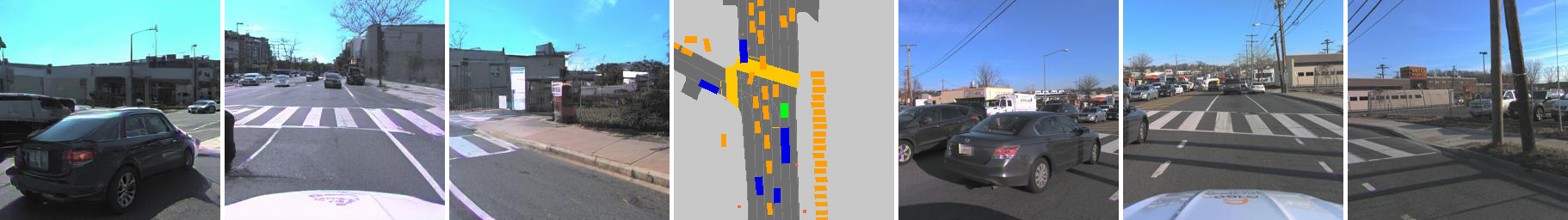}
    \vfil
   \centering
   \vspace{0.05cm}
   \includegraphics[width=1.0\linewidth]{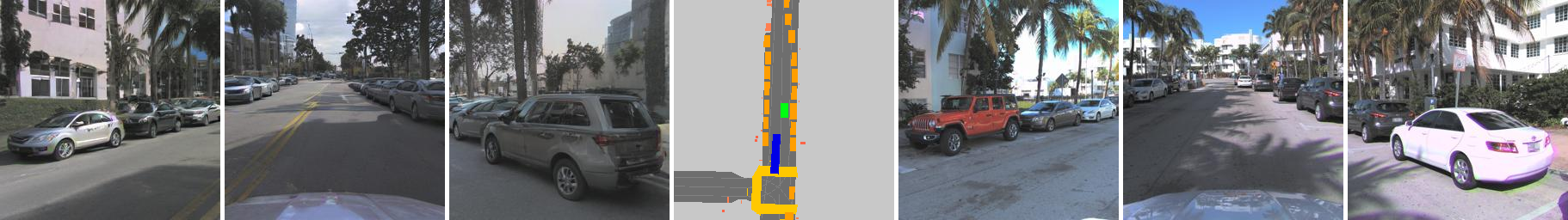}
   \caption{Comparison to real-views on Argoverse 2. Images on the left are synthesized, images on the right are corresponding real ones.}
   \label{fig:argoverse_gen}
\end{figure*}

In~\cref{fig:argoverse_gen}, we compare synthesized images to real ones for the same BEV layouts on Argoverse 2. We observe that our model places vehicles in the same location as the real one, even when a vehicle is in multiple views, demonstrating that the model learns to synthesize coherent content across views.

\begin{figure}[t]
  \centering
   \includegraphics[width=1.0\columnwidth]{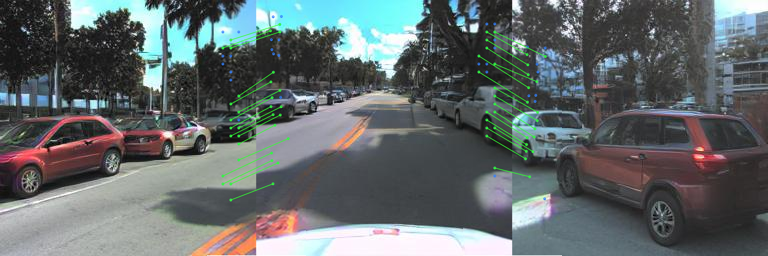}\vfil
   \includegraphics[width=1.0\columnwidth]{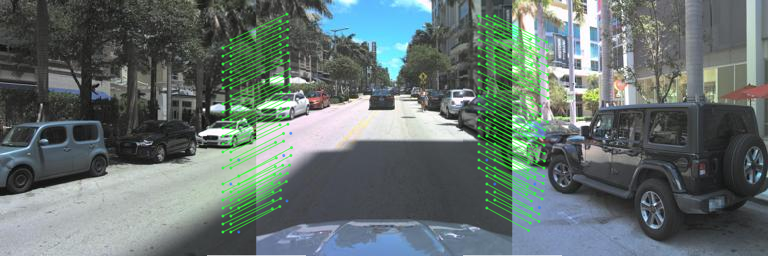}
   \caption{We display point correspondences (blue) obtained with LoFTR for the VSC metric. Green lines indicate inliers from RANSAC. Top: Generated images, Bottom: Real images.}
   \label{fig:vsc}
\end{figure}

\paragraph{Quantitative result} 
\label{ssec:quantitative}
We use the Fréchet Inception Distance (FID) to evaluate our synthesized quality compared to the source images. Metrics are calculated on the validation set for each respective dataset. We employ no post-generation filtering. For calculating FID scores, we use clean-fid~\cite{parmar2021cleanfid}.

To differentiate between the performance of our 1st and 2nd stage, we compare our results to the results obtained by feeding the encoded tokens directly to the decoder, as is done when training the 1st stage. This represents the theoretical upper bound of our model's performance, largely removing the effect of the 1st stage which is not the focus of this paper.

As seen in \cref{tab:baselines}, our \textbf{BEVGen} model achieved an FID score of $25.54$ on nuScenes, compared to the baseline score of $138.30$. This is in comparison to our reference upper-bound FID score of $9.37$. Our model utilizing our sparse masking design from Sec. 3.4 achieved an FID score of $28.67$. This sparse variant is approximately 48\% faster during inference and 40\% faster for training. On Argoverse 2, our model achieves an FID score of $25.51$.

While FID is a common metric to measure image synthesis quality, it fails to entirely capture the design goals of our task and cannot reflect the synthesis quality of different semantic categories. Since we seek to generate multi-view images consistent with a BEV layout, we wish to measure our performance on this consistency. To do this, we leverage a pre-trained BEV segmentation network CVT from~\cite{zhouCrossviewTransformersRealtime2022}. We apply the CVT to the generated images and then compare the predicted layout with the ground-truth BEV layout. We report both the road and vehicle class mean intersection-over-union (mIoU) scores. As shown in \cref{tab:baselines}, we beat our baseline by $4.4$ and $1.45$ for road and vehicle classes respectively. Note that the performance of the BEV segmentation model on the validation set is $66.31$ and $27.51$ for road and vehicle mIoU respectively. This reveals that though the model can generate road regions in the image in a reasonable manner, it still has a limited capability of generating high-quality individual vehicles that can be recognized correctly by the segmentation network. This is a common problem for scene generation where it remains challenging to synthesize the small objects entirely. Our work is a starting point and we plan to improve small object synthesis in future work.

Finally, we seek to quantitatively verify the cross-view consistency of our model. To do this, we introduce a consistency metric that looks at the similarity between overlapping portions of images. For example, for the front-left and front-center camera on nuScenes, we extract a window on the right and left edges of the images, respectively. We then attempt to find keypoint correspondences between these image patches by performing feature matching with~\cite{sunLoFTRDetectorFreeLocal2021}. As each correspondence has a confidence, we simply find the sum of these confidences and average across all windows to obtain a metric for cross-view consistency. We call this metric the View Consistency Score (VSC). On nuScenes, synthesized view images from our model obtain a VSC of $5.65$, with the corresponding real images obtaining a result of $12.86$. On Argoverse 2, synthesized view images obtain a VSC of $13.00$, with the corresponding real images obtaining a result of $42.90$. We see further results in~\cref{tab:ablations}. We show matched keypoints in~\cref{fig:vsc}, with both images corresponding to the same BEV layout. We note that this metric does not evaluate whether there is a proper perspective shift between the viewpoints, only that there are point correspondences in overlapping regions. Qualitatively, BEVGen consistently generates a realistic shift in perspective between views.

As far as we know, our work is the first attempt at synthesizing street views conditioned on a BEV layout. To establish a comparison baseline, we reproduce one of the cross-view synthesis methods proposed in \cite{regmiCrossViewImageSynthesis2018}. Firstly, we project the BEV semantic mask onto the camera perspective view using extrinsics and intrinsics. Next, we utilize a conditional GAN to generate a street view semantic mask from this warped BEV layout. Finally, we use another conditional GAN to perform the semantic mask-to-image translation. As neither the nuScenes nor the Argoverse datasets provide 2D segmentation labels, we use a pre-trained segmentation model \cite{yuanObjectContextualRepresentationsSemantic2020} to generate pseudo ground-truth labels for the first conditional GAN. Following~\cite{regmiCrossViewImageSynthesis2018}, we use the Pix2Pix~\cite{isolaImageToImageTranslationConditional2017} model for the first and second stage. However, at their core, existing cross-view synthesis methods such as this one are not designed to handle the BEV generation task as, at their core, they translate existing rich RGB information rather than generate new images from a sparse semantic layout.

\subsection{Ablation Study}
\label{ssec:ablation}
To verify the effectiveness of our design choices, we run an ablation study on key features of our model. We run these experiments on the same subset of the nuScenes validation set as in \cref{ssec:results}, but only consider the 3 front-facing views to reduce training time. The 3 front-facing views have a larger FoV overlap and capture more relevant scene features compared to the side-facing rear views. This area is more relevant to our task and still allows us verify the model design objectives.

We test four variants of our model, one with only center-out decoding, one with our camera bias, one with the camera bias and spatial embeddings, and a final model that we train from scratch using our sparse masking, instead of fine-tuning. \cref{tab:ablations} shows a steady decrease in FID scores and increase in VSC, as we add the camera bias, and spatial embeddings.

\begin{table}
  \centering
  \setlength{\tabcolsep}{3.85pt}
  \begin{tabular}{@{}lcc@{}}
    \toprule
    Method & FID$\downarrow$ & VSC$\uparrow$\\ \midrule
    Row-major Decoding& 43.18 & 4.33\\
    Center-out Decoding& 42.32 & 4.53 \\     
    $+$ Camera Bias & 41.20 & 4.74\\    
    $+$ Camera Bias, Spatial Embedding & \textbf{40.48} & \textbf{6.24} \\
    $+$ Sparse, Camera Bias, Spatial Embedding & 48.31 & 5.44 \\
    \bottomrule
  \end{tabular}
  \caption{Ablation of the key model components.}
  \label{tab:ablations}
\end{table}

\section{Applications}
\label{sec:applications}
Generating realistic images from BEV layout has many applications. In this section we explore the applications of data augmentation for BEV segmentation and image generation from simulated BEV.

\begin{table*}
  \centering
    \begin{tabular}{@{}lcccccc@{}}
      \toprule
      & \multicolumn{2}{c}{CVT} & \multicolumn{2}{c}{BEVFormer Tiny} & \multicolumn{2}{c}{BEVFormer Small} \\
      & Road mIoU$\uparrow$ & Vehicle mIoU$\uparrow$ & mAP$\uparrow$ & NDS$\uparrow$ & mAP$\uparrow$ & NDS$\uparrow$ \\
      \midrule
      w/o augmentation & 71.3 & 36.0 & 25.7 & 35.9 & 37.0 & 47.9\\
      w/ augmentation & \textbf{71.9} & \textbf{36.6} & \textbf{27.3} & \textbf{37.2} & \textbf{39.9} & \textbf{50.6}\\
      \bottomrule
    \end{tabular}
  \caption{Data augmentation results for BEV Segmentation and Detection on the validation set of nuScenes.}
  \label{tab:data_aug}
\end{table*}

\begin{figure*}[th]
  \centering
   \includegraphics[width=0.495\linewidth]{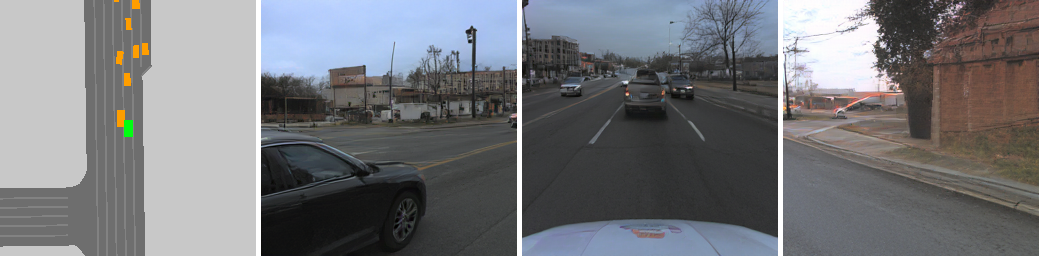}\hfil
   \includegraphics[width=0.495\textwidth]{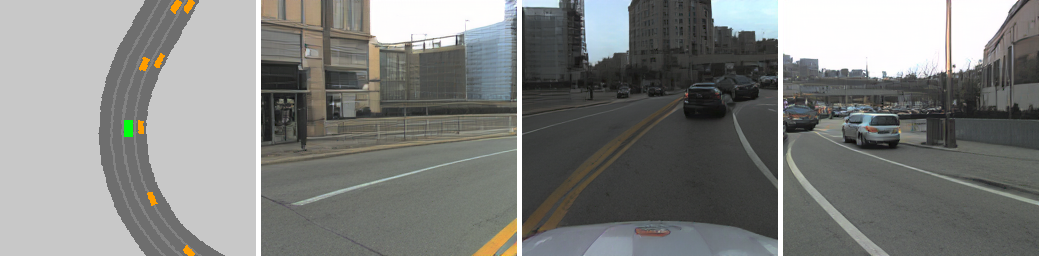}
   \vfil
   \centering
   \vspace{0.05cm}
   \includegraphics[width=0.495\linewidth]{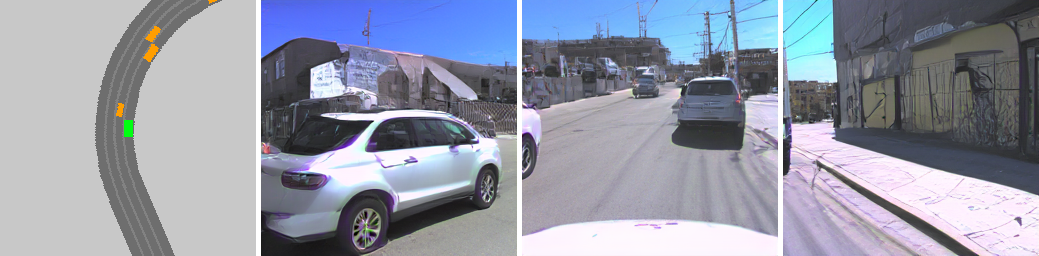}\hfil
   \includegraphics[width=0.495\textwidth]{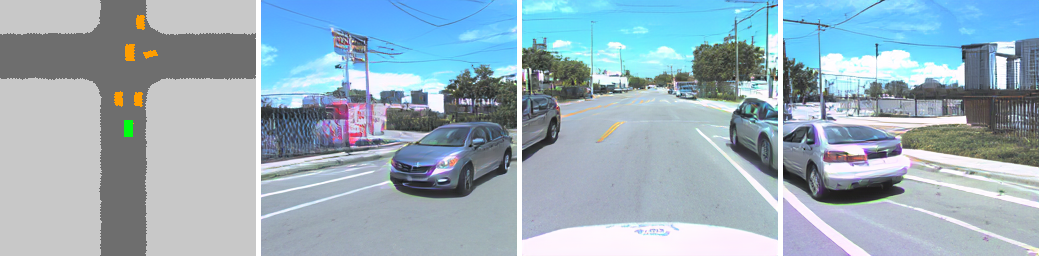}\hfil
   \caption{Generating images based on the BEV layouts provided by the MetaDrive driving simulator}
   \label{fig:metadrive}
\end{figure*}

\begin{figure}[th]
  \centering
   \includegraphics[width=1.0\columnwidth]{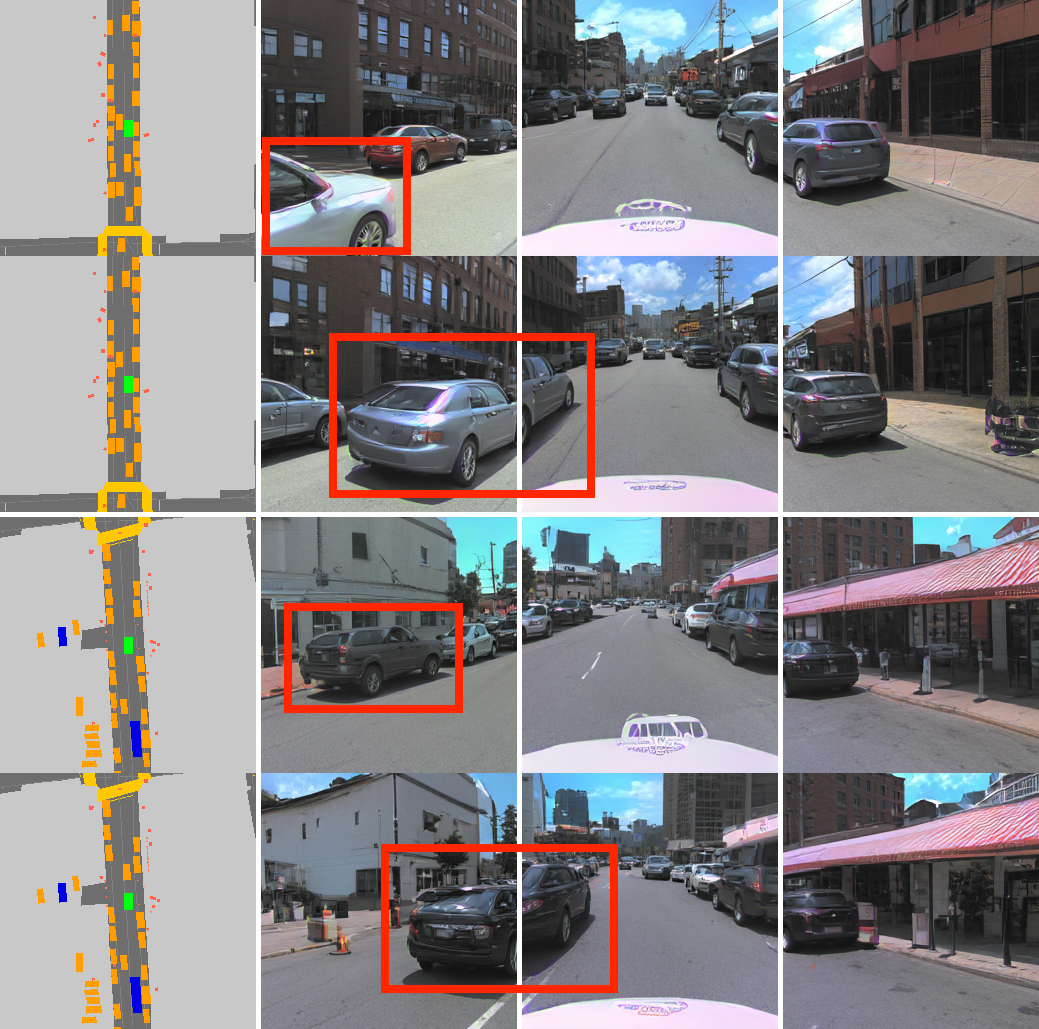}
   \caption{We modify two existing BEV layouts to demonstrate spatial disentanglement of the model.}
   \label{fig:modified}
\end{figure}

\paragraph{Image generation from simulated BEV}
Since one motivation for our task definition lies in the simplicity of the BEV layout, we wish to determine whether this enables our model to generate new scenes from out-of-domain (OOD) HD maps. We use MetaDrive simulator~\cite{li2022metadrive} to procedurally generate traffic scenarios and input these BEV layouts into \textbf{BEVGen}. Generated images are shown in \cref{fig:metadrive}. Scenario generation that addresses the long-tail problem is itself an active area of research~\cite{fengTrafficGenLearningGenerate2023b,liScenarioNetOpenSourcePlatform2023} and outside the scope of our work, but we clearly demonstrate that we can synthesize such synthetic scenarios into real images using the BEV layout as a bridge. This has potential to address the sim2real gap.

Additionally, we demonstrate that we can generate a safety-critical scenario by editing an existing scene. This functionality is an important component of the safety testing for many autonomous driving systems as it allows for both permutation testing for real-life scenarios but also for a human developer to construct difficult scenarios would otherwise be dangerous and observe the behavior of a system. We demonstrate this editing ability in~\cref{fig:modified} where we translate cars in two different scenes.

\paragraph{Data augmentation for BEV segmentation}
\label{ssec:augmentation}
An important application of our BEV conditional generative model is generating synthetic data to improve prediction models. Thus, we seek to verify the effectiveness of our model by incorporating our generated images as augmented samples during the training of multiple camera-only BEV segmentation and 3D detection models. For BEV Segmentation, we use CVT~\cite{zhouCrossviewTransformersRealtime2022}, which is also used in \cref{ssec:results}, and compare our results to training without any synthetic samples. We generate 6,000 unique instances using the BEV layout from the \emph{train} set on nuScenes and use these to augment the original training set. These synthetic instances are associated with the ground truth BEV layout for training, with no relation to results from \cref{ssec:results}. For the detection task, we verify on the state-of-the-art detector BEVFormer~\cite{liBEVFormerLearningBird2022a}. We select two model sizes for testing and apply the same augmentation regime used for CVT. As seen in~\cref{tab:data_aug}, our synthetic data improves validation mIoU for BEV Segmentation by $0.6$ for both the road category and the vehicle category. Additionally, we see more substantial gains in BEV Object Detection with improvements of $1.6$ and $1.3$ for mAP and NDS respectively for the \texttt{Tiny} model and $2.9$ and $2.7$ respectively for the \texttt{Small} model.
\section{Conclusion}
\label{sec:conclusion}
In this work we introduce this new BEV generation task and propose and approach in the form of a generative model we call \textbf{BEVGen}. After training on real-world driving datasets, the proposed model can generate spatially consistent multi-view images from a given BEV layout. We further show its application in data augmentation and simulated BEV generation.

\section{Acknowledgments}
This work was supported by the National Science Foundation under Grant No. 2235012.

{\small
\bibliographystyle{IEEEtran}
\bibliography{IEEEabrv,citations}
}

\end{document}